\documentclass[11pt]{article}

\usepackage[final]{acl}
\makeatletter
\acl@anonymizefalse
\makeatother
\usepackage[english]{babel}
\usepackage{fontspec}
\usepackage{xeCJK}
\xeCJKsetup{CJKspace=true}
\setCJKmonofont{FandolSong-Regular.otf}[Script=Default]

\usepackage{amsmath}
\usepackage{graphicx}
\usepackage{booktabs}
\usepackage{multirow}
\usepackage{tabularx}
\usepackage{float}
\usepackage{placeins}
\usepackage{xcolor}
\usepackage{microtype}
\definecolor{routeblue}{HTML}{245E74}

\newcommand{\method}{\textsc{LoopRoute}}
\newcommand{\caseicon}[1]{\raisebox{-0.16em}{\includegraphics[height=1.25em]{#1}}}
\newcommand{\llmlogo}[1]{\raisebox{-0.12em}{\includegraphics[height=0.85em,width=1.1em,keepaspectratio]{#1}}\nobreak\hspace{0.2em}}

\title{Stop Going in Circles:Loop-Triggered Routing for System~1 Decision Agents}

\author{
Zeming yu\textsuperscript{1 }\thanks{These authors contributed equally to this work.},
Huamin Ma\textsuperscript{1 *},
Yingyu Chen\textsuperscript{1}, Yuan Sun\textsuperscript{1}, Peng Hu\textsuperscript{1}\thanks{Corresponding author.}, Yang Qin\textsuperscript{2†} \\
  \textsuperscript{1}College of Computer Science, Sichuan University\\
  \textsuperscript{2}Huawei Technologies \\
  {\small\texttt{\{yuzeming18, huaminma2006, cyy262511, penghu.ml, qinyang.gm\}@gmail.com}, \texttt{sunyuan\_work@163.com}}
}

\begin{document}
\maketitle

\begin{abstract}
System~1 decision models (represented by Jev)  provide fast, low-cost action selection, but existing agent integrations do not reliably translate these local advantages into task completion. At certain points during execution, an agent may repeatedly select the same actions without making progress toward its goal. 
LLM-based policies offer strong reasoning capabilities that support planning within an agent, but invoking them at every step incurs substantial overhead. The effective coordination should therefore preserve Jev's fast local action selection while introducing LLM reasoning when execution stalls. We observe that recurring state–action pairs can signal stalled execution. 
We therefore propose LoopRoute, a simple but effective action router that uses this signal to allocate control within an agent. 
The decision model acts by default. When a loop is detected, the LLM provides brief, feedback-driven assistance and then returns control to the decision model. Later loops can request further assistance. The rule requires neither additional training nor model-specific confidence calibration. Across three benchmarks and seven LLMs, LoopRoute improves task success over standalone Jev with only limited use of the LLM policy. The mean gains are 21.0\%, 60.1\%, and 30.1\% on ALFWorld, MapReader, and TextWorld cooking, respectively. Compared with full LLM control, LoopRoute also substantially reduces the mean cumulative API time and API-call cost per task. The same rule also improves the success rates of six frozen decision models. 

\end{abstract}

\section{Introduction}
\label{sec:intro}
\begin{figure}[!t]
\centering
\includegraphics[width=\columnwidth,height=8.8cm,keepaspectratio]{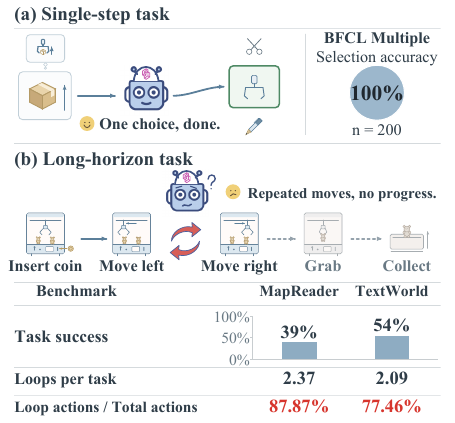}
\caption{Local selection versus long-horizon execution with standalone Jev (robot). (a) Static tool selection on BFCL Multiple. (b) Illustrative loops stall a multi-step task. Loop onsets are averaged over all tasks. Repeated-step shares are pooled over failed tasks only. TextWorld denotes cooking (Appendix~\ref{app:diagnostics}).}
\label{fig:motivation}
\end{figure}

Completing long-horizon tasks~\citep{yao2023react,ALFWorld20} requires interactive agents to execute sequences of actions in dynamic environments. We observe that Jev performs well in next-action selection when the goal is clear. It performs poorly, however, when action selection requires planning and reasoning. Further experiments
show that System~1 decision models alone cannot overcome stagnation during long-horizon execution. Invoking an LLM at every step, however, incurs substantial inference overhead. We therefore seek a strategy that coordinates the efficient, low-cost action selection of System~1 decision models such as Jev~\citep{almeida2026jev,arxiv261017205,arxiv261021244} with the stronger task-level reasoning and recovery capabilities of large language models (LLMs)~\citep{yao2023react,rozanov2025stateact}. The goal is to improve action-selection efficiency while maintaining a high task success rate.

Existing research has explored this problem through LLM-based reasoning, the fast-slow policy collaboration, and execution-feedback-driven decision switching~\citep{yao2023react,arxiv230517390,arxiv250805081}. LLM-based approaches use reasoning and reflection to handle complex tasks, but repeated calls increase execution costs. The fast-slow methods combine the advantages of the two decision mechanisms. For example, SwiftSage requires training a fast policy, while JevAgentHarness calibrates confidence thresholds
for individual decision models~\citep{arxiv230517390,arxiv261021412}. These requirements hinder reliable transfer across decision models or tasks. Execution-feedback-driven methods adjust decision policies when progress stalls or failures recur~\citep{arxiv250805081}. Moreover, detecting a stall does not ensure that the next policy will choose an action that advances the task~\citep{arxiv261009683}. 
An attempt first tests candidate actions in a simulator and then asks Jev to choose using the simulated outcomes. This
approach is difficult to apply when action outcomes cannot be simulated in advance~\citep{arxiv261001834}. Given these limitations, we seek a simple, shared execution signal that indicates when System~1 decision models need additional reasoning. The signal should require neither additional training nor model-specific confidence calibration. However, when calibrated confidence is unreliable, determining when an LLM should intervene remains a challenge.

To address this problem, we propose \method{}, a dynamic action routing mechanism triggered by recurring state--action pairs. Its central idea is to use the agent's execution history to identify potential decision stagnation. When a fast decision model repeatedly selects the same action under the same observed state, the system treats this behavior as a signal for additional reasoning and temporarily transfers control to an LLM. Specifically, \method{} uses a frozen System~1 model as the default controller. Once repeated behavior is detected, the LLM selects alternative actions based on the task goal, current observation, and execution history. After a bounded takeover period, control is returned to the original decision model. In this way, the LLM no longer needs to participate continuously in every action-selection step, but instead provides on-demand reasoning assistance when local decisions become stagnant. Furthermore, since the routing signal is derived from shared execution history, the mechanism requires neither additional training, model-specific confidence calibration, nor simulated action outcomes, making it applicable to different frozen decision models.

We evaluate \method{} on three interactive benchmarks, ALFWorld, MapReader, and TextWorld cooking, using seven LLM policies. Experimental results show that \method{} improves the long-horizon task performance of fast decision models while reducing execution time and monetary cost compared with continuous LLM control. Further threshold analysis, controlled experiments, and cross-model transfer evaluations validate the effectiveness and generality of the routing mechanism. We also investigate the limitations of repetition-triggered recovery in the presence of irreversible actions.

Our contributions are summarized as follows:
\begin{itemize}
    \item We propose \method{}, a dynamic routing mechanism that uses recurring state--action pairs to identify execution stagnation and enable on-demand collaboration between frozen System~1 decision models and LLMs, without additional training or model-specific confidence calibration.

    \item We evaluate the effectiveness and efficiency of \method{} across three interactive benchmarks and seven LLM policies, and investigate the role of the routing mechanism through threshold analysis and controlled experiments.

    \item We investigate the transferability and limitations of the proposed mechanism by evaluating the unchanged routing rule across six frozen decision models and analyzing the impact of irreversible actions on task recovery.
\end{itemize}

\section{Related Work}
\label{sec:related}

\paragraph{System~1 decision models.}
System~1 decision models score caller-defined options without generating a reasoning trace, extending earlier work on ranking admissible commands and skills~\citep{he2016drrn,ichter2023saycan}. Recent models span three broad groups. \emph{Hosted general-purpose models} include Jev~\citep{almeida2026jev}, OpenAI's Decisions API~\citep{openai2026decisions}, Microsoft-Decision-1~\citep{microsoft2026decision1}, and Liquid AI's d1~\citep{liquid2026d1}. These models expose structured decisions as a service. The d1 model also accepts images. \emph{Open-weight models} make local deployment and adaptation possible: Cloudflare's Clef and Clef-flash~\citep{cloudflare2026clef}, Perplexity's pplx-decider~\citep{perplexity2026decider}, Strands Decider~\citep{strands2026decider}, and Together AI's Tev1~\citep{together2026tev1} provide general decision models and training recipes. This group also includes compact models such as StartLux-4B~\citep{startlux2026decision}, Kev-4B~\citep{palmer2026kev4b}, and Laya~\citep{laya2026}, as well as multimodal models such as OneJev~\citep{onejev2026}, JEV-27B-VL~\citep{autotrust2026jevvl}, and Intern-Decision~\citep{arxiv261022131}. \emph{Domain-specific models} adapt decision prediction to a particular workflow. For example, SearchJev targets relevance, evidence sufficiency, and action selection in search agents~\citep{arxiv261005107}.

Open training recipes~\citep{arxiv261017205} and comparative benchmarks~\citep{arxiv261021244,arxiv261002267} demonstrate growing interest in efficient decision prediction. Most evaluations, however, assess individual decisions. Studies of complete agent workflows find that lower decision cost does not consistently improve task outcomes, with stalled execution and failed recovery remaining obstacles~\citep{arxiv261013092,arxiv261001834}. We study how execution feedback can coordinate these models' local selection strengths with LLM assistance for task completion.

\paragraph{Coordination and recovery in language agents.}
Fast--slow agents combine direct action prediction with LLM reasoning for planning and recovery. SwiftSage pairs a trained fast policy with a planner~\citep{arxiv230517390}. CogniWeb switches between direct and deliberative modes after repeated failures~\citep{arxiv250805081}, while System Switch combines typed decisions with a reasoning module and execution gates~\citep{arxiv261009683}. Model cascades and learned routers balance quality and cost~\citep{arxiv230505176,arxiv240618665}. At the agent level, routing can use accumulated experience~\citep{arxiv260102695} or uncertainty-based deferral~\citep{arxiv260407036,arxiv261021412}.

Execution feedback provides another basis for coordination. Repeated action patterns are a common failure mode in language agents~\citep{arxiv261019376}. Reflexion uses action--response repetition as a cue for reflection~\citep{arxiv230311366}. DeLoop trains agents to prefer later non-looping actions~\citep{arxiv261019376}, and the Loop Breaker in VLAA-GUI forces a strategy change when screen states recur~\citep{arxiv260421375}. We use recurring state--action pairs to route control from a frozen decision model to an LLM policy for brief assistance. Our controls distinguish suppressing repetition from selecting a productive alternative (\S\ref{sec:exp-break}). The transfer evaluation tests whether the same rule remains effective across decision models.

\section{\method{}}
\label{sec:method}
\begin{figure*}[t]
\centering
\includegraphics[width=\textwidth]{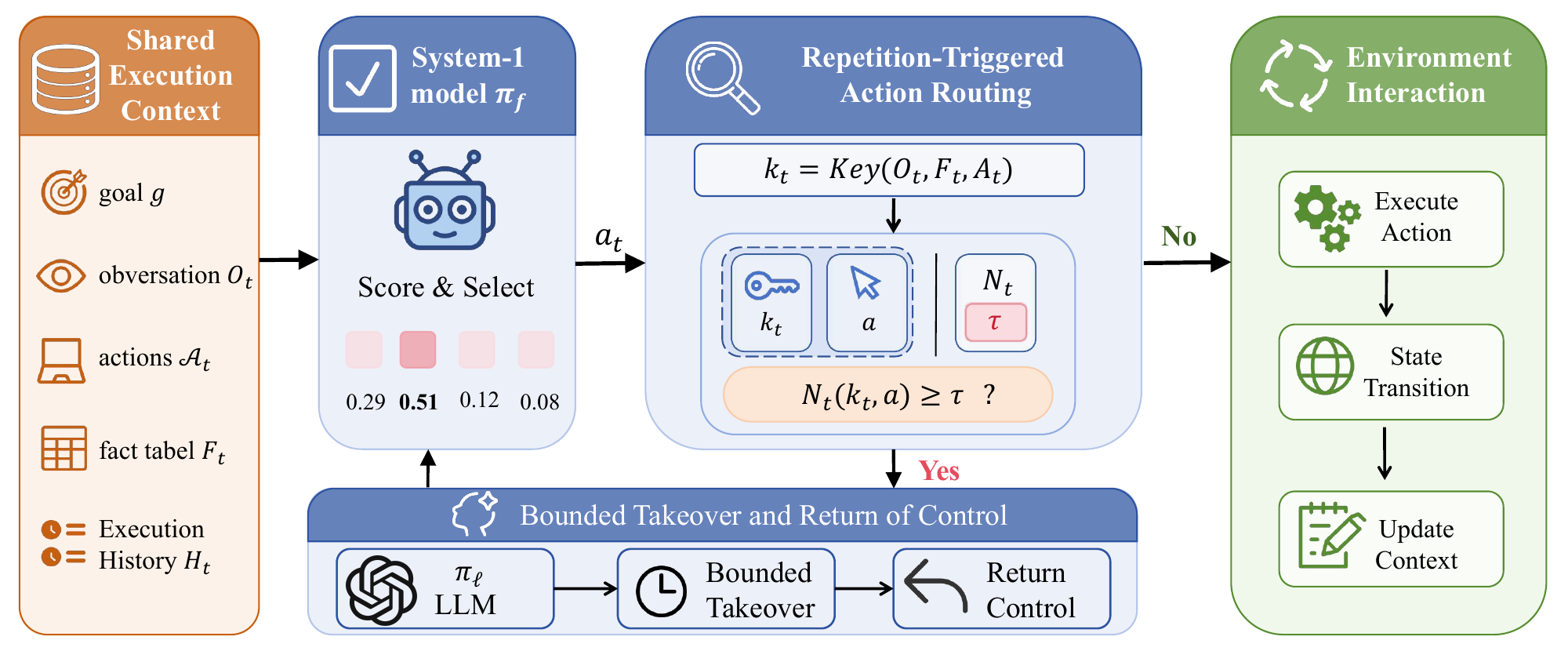}
\caption{Overview of \method{}. The System~1 decision model proposes an action from a shared execution context. The router checks how often the proposed state--action pair has been executed: proposals below the repetition threshold execute directly, while a repeated proposal triggers at most two feedback-driven LLM actions before control returns to the decision model. Both policies update the same execution context after each environment transition.}
\label{fig:overview}
\end{figure*}

\method{} coordinates local action selection and task-level recovery through the execution sequence in Figure~\ref{fig:overview}. We first organize task evidence for the default Jev policy, then monitor whether its actions revisit the same observed situation. Such recurrence triggers brief LLM assistance. A shared context carries the resulting feedback across each transfer of control.

\subsection{Structured Context and Action Selection}
\label{sec:setting}
A local decision must relate the latest observation to the task goal and evidence from earlier interactions. We organize these inputs as a structured execution context. At step $t$,
\begin{equation}
x_t = (g, o_t, \mathcal{A}_t, F_t, H_t).
\label{eq:context}
\end{equation}
The goal $g$ specifies the desired outcome, and $o_t$ supplies current feedback. The admissible set $\mathcal{A}_t$ grounds selection in executable actions. The confirmed-fact table $F_t$ preserves discoveries, such as object locations and verified conditions, beyond the current observation. The history $H_t$ retains the order of executed actions and responses. Together, these fields relate the available choices to the goal and execution so far. Benchmark-specific construction is given in Appendix~\ref{app:impl}.

Our comparison of input contexts supports this enriched input over the goal and observation alone (Appendix~\ref{app:impl}). Jev uses $x_t$ to score admissible actions in one forward pass and select $\hat{a}_t=\pi_f(x_t)$ under its frozen policy $\pi_f$. This preparation supports task-directed choices, but repeated selections can still leave execution at the same obstacle.

\subsection{Repetition-Triggered Action Routing}
\label{sec:when}
Structured input supports action selection, but does not itself identify stalled progress. A repeated action can be useful after the state changes. We therefore detect recurrence of state--action pairs rather than actions alone.

\paragraph{State keys.}
We derive a comparable signature from the public state fields in the context:
\begin{equation}
k_t = \mathrm{Key}(o_t, F_t, \mathcal{A}_t)
\label{eq:key}
\end{equation}
The key uses benchmark-specific public fields, such as location, inventory, verified object states, and admissible actions (Appendix~\ref{app:impl}). We exclude the growing history and step index, which change even in a loop. Recurrence is thus checked against the selected observed fields rather than the complete hidden state.

\paragraph{Interception before execution.}
Given Jev's proposal, let $N_t(k,a)$ be the retained count of executions of action $a$ under key $k$. The router checks
\begin{equation}
r_t = \mathbf{1}\!\left[N_t(k_t,\hat{a}_t)\geq\tau\right],\quad \tau=2.
\label{eq:trigger}
\end{equation}
When the same pair has already been executed twice, we treat $r_t=1$ as a signal of stalled execution. The router intercepts the next proposal and transfers control to the LLM. The intercepted proposal consumes no environment step and is not added to the history. Otherwise, Jev's action executes and the records update. The check uses execution counts rather than model-specific confidence scores, with no extra model call. Counter updates and resets are specified in Appendix~\ref{app:impl}, and \S\ref{sec:exp-threshold} examines the shared threshold. A confidence-based comparison is in Appendix~\ref{app:conf}.

\subsection{Bounded Takeover and Return of Control}
\label{sec:how}
Recurrence signals stalled execution but cannot select a productive alternative. The LLM policy $\pi_\ell$ reasons about the remaining task from $x_t$ to choose an action. ReAct and StateAct prompts instantiate this policy. Fresh environment feedback informs each subsequent decision.

To advance stalled execution without unnecessarily replacing Jev during productive local decisions, each LLM takeover is bounded by $B=2$ actions. Let $b_t$ be the remaining budget, initially zero. The router evaluates $r_t$ under Jev control and sets $r_t=0$ during a takeover. The executed action is
\begin{equation}
a_t=\begin{cases}
\pi_\ell(x_t), & b_t>0\ \text{or}\ r_t=1,\\
\hat{a}_t, & \text{otherwise}.
\end{cases}
\label{eq:routed-action}
\end{equation}
After a nonterminal transition, the budget updates as
\begin{equation}
b_{t+1}=\begin{cases}
b_t-1, & b_t>0,\\
B-1, & b_t=0\ \text{and}\ r_t=1,\\
0, & \text{otherwise},
\end{cases}
\label{eq:takeover-budget}
\end{equation}
The first LLM action consumes one step of this budget. If the environment terminates, the takeover ends immediately.

\paragraph{Returning control and intervening again.}
When the budget reaches zero, Jev resumes from the updated context without a separate recovery test. Later recurrence can trigger another bounded takeover. This cycle retains fast local execution between interventions and recruits further assistance when progress stalls.

\subsection{Shared Execution Context}
\label{sec:share}
To carry the effects of assistance into later Jev decisions, both policies read and update the same history and confirmed-fact table. After action $a_t$ and observation $o_{t+1}$,
\begin{equation}
\begin{aligned}
H_{t+1}&=H_t\mathbin{\oplus}(a_t,o_{t+1}),\\
F_{t+1}&=\mathcal{U}(F_t,a_t,o_{t+1}),
\end{aligned}
\label{eq:shared-update}
\end{equation}
Here $\oplus$ appends the executed action and response, and $\mathcal{U}$ records only facts confirmed by the environment. These updates apply under either controller. The LLM receives Jev's preceding execution, and Jev resumes with the evidence produced during assistance. All decision-model baselines receive the same structured context, so their comparison with \method{} evaluates routing on top of this input preparation. Table~\ref{table:trajectory} illustrates the resulting cycle from stalled selection through LLM assistance to resumed Jev execution.

\begin{table}[t]
\centering
\fontsize{10}{12}\selectfont
\setlength{\tabcolsep}{3pt}
\begin{tabularx}{\columnwidth}{@{}rl>{\raggedright\arraybackslash}X@{}}
\multicolumn{3}{@{}l}{\textit{Goal: put a cool tomato in microwave.}} \\
\toprule
Step & Actor & Action $\rightarrow$ observation \\
\midrule
23 & \caseicon{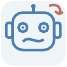} Jev & examine microwave 1 $\rightarrow$ \textit{closed} \\
24 & Jev & open microwave 1 $\rightarrow$ \textit{you see a tomato 1} \\
25 & Jev & close microwave 1 \\
26 & Jev & examine microwave 1 $\rightarrow$ \textit{closed} \\
27 & Jev & \textcolor{gray}{Proposes \textit{examine microwave 1}: intercepted} \\
\midrule
27 & \caseicon{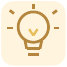} \textbf{LLM} & open microwave 1 $\rightarrow$ \textit{you see a tomato 1} \\
28 & \textbf{LLM} & take tomato 1 from microwave 1 \\
\midrule
29--31 & \caseicon{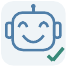} Jev & \dots{} go to fridge 1, cool tomato 1 with fridge 1 \\
32--33 & Jev & go to microwave 1, move tomato 1 to microwave 1 \\
\bottomrule
\end{tabularx}
\caption{An ALFWorld trajectory in which Jev cycles through microwave actions after storing an uncooled tomato. \method{} intercepts its third proposal to \textit{examine microwave 1} under the same observed state. The LLM policy takes two actions and returns control. Jev completes the task at step~34. Jev alone reaches the 100-step limit. The icons mark stalled selection, LLM assistance, and resumed progress.}
\label{table:trajectory}
\end{table}

\section{Experiments}
\label{sec:exp}
We evaluate whether loop-triggered coordination converts Jev's fast local selection into task-level gains. The main results jointly assess success, per-task time, and cost, including controls without LLM assistance. Further analyses examine the loop signal and the timing and duration of takeovers. The transfer experiments reuse the same rule across frozen decision models.

\subsection{Experimental Setup}
\label{sec:setup}
\paragraph{Benchmarks.}
We choose three text environments that share an admissible-action interface, require sustained multi-step interaction, and permit evaluation by task completion. Their common interface supports direct action selection by both decision models and LLM policies, while their tasks test different execution demands: household manipulation in ALFWorld~\citep{ALFWorld20} (134 \texttt{valid\_unseen} tasks), navigation and retrieval in TextWorldExpress MapReader~\citep{jansen2023textworldexpress} (100 tasks), and ingredient preparation with irreversible actions in TextWorld cooking~\citep{cote18textworld} (100 games, 20 per difficulty level from 5 to 9). Each episode allows at most 100 actions.

\begin{table*}[t]
\centering
\fontsize{9}{10.8}\selectfont
\setlength{\tabcolsep}{2.2pt}
\begin{tabular*}{\textwidth}{@{\extracolsep{\fill}}ll*{9}{r}@{}}
\toprule
& & \multicolumn{3}{c}{ALFWorld} & \multicolumn{3}{c}{MapReader} & \multicolumn{3}{c}{TextWorld} \\
\cmidrule(lr){3-5}\cmidrule(lr){6-8}\cmidrule(l){9-11}
Model & Policy & SR (\%) & Time (s) & Cost (\$) & SR (\%) & Time (s) & Cost (\$) & SR (\%) & Time (s) & Cost (\$) \\
\midrule
Jev & Alone & 77.6 & 13.4 & 0.0074 & 39.0 & 20.9 & 0.0059 & 54.0 & 16.5 & 0.0096 \\
Jev & + Rank & 88.8 & 13.4 & 0.0062 & 62.0 & 16.4 & 0.0050 & 66.0 & 13.0 & 0.0081 \\
Jev & + Requery & 82.8 & 15.3 & 0.0077 & 62.0 & 22.4 & 0.0072 & 66.0 & 17.0 & 0.0116 \\
\midrule
GPT-6-Astra & Full & 100.0 & 64.7 & 0.4600 & 100.0 & 161.6 & 0.4582 & 100.0 & 193.8 & 0.7517 \\
 & Routed & 100.0 & 18.5 & 0.0528 & 100.0 & 85.1 & 0.1771 & 83.0 & 79.8 & 0.2700 \\
\midrule
Claude Opus 5 & Full & 100.0 & 61.0 & 0.3090 & 100.0 & 109.2 & 0.2935 & 91.0 & 141.5 & 0.6622 \\
 & Routed & 98.5 & 18.0 & 0.0511 & 99.0 & 70.5 & 0.1896 & 81.0 & 64.1 & 0.3271 \\
\midrule
Gemini 3.8 Flash & Full & 100.0 & 238.0 & 0.0243 & 100.0 & 316.0 & 0.0244 & 100.0 & 427.2 & 0.0487 \\
 & Routed & 100.0 & 35.5 & 0.0078 & 100.0 & 133.5 & 0.0133 & 83.0 & 137.2 & 0.0235 \\
\midrule
Grok 4.5 & Full & 100.0 & 155.5 & 0.0883 & 100.0 & 168.5 & 0.0889 & 100.0 & 161.8 & 0.1368 \\
 & Routed & 100.0 & 30.0 & 0.0138 & 100.0 & 114.4 & 0.0464 & 87.0 & 65.1 & 0.0547 \\
\midrule
GLM-5.3 & Full & 100.0 & 19.2 & 0.0274 & 100.0 & 28.8 & 0.0243 & 95.0 & 59.4 & 0.0570 \\
 & Routed & 99.3 & 15.9 & 0.0098 & 100.0 & 23.0 & 0.0143 & 89.0 & 41.7 & 0.0382 \\
\midrule
Qwen3.8-Flash & Full & 97.8 & 27.4 & 0.0092 & 100.0 & 59.3 & 0.0038 & 82.0 & 139.7 & 0.0160 \\
 & Routed & 97.8 & 11.5 & 0.0055 & 98.0 & 73.9 & 0.0055 & 83.0 & 71.2 & 0.0124 \\
\midrule
DeepSeek-V4.1-Flash & Full & 94.0 & 157.3 & 0.0237 & 98.0 & 193.0 & 0.0118 & 92.0 & 198.0 & 0.0219 \\
 & Routed & 94.8 & 23.6 & 0.0074 & 97.0 & 102.4 & 0.0083 & 83.0 & 92.4 & 0.0131 \\
\midrule
\multicolumn{2}{@{}l}{\textbf{Mean routed}} & \textbf{98.6} & \textbf{21.9} & \textbf{0.0212} & \textbf{99.1} & \textbf{86.1} & \textbf{0.0649} & 84.1 & \textbf{78.8} & \textbf{0.1056} \\
\multicolumn{2}{@{}l}{\textbf{Mean gain vs. Full}} & \textbf{-0.2\%} & \textbf{66.8\%}$\downarrow$ & \textbf{71.0\%}$\downarrow$ & \textbf{-0.6\%} & \textbf{30.7\%}$\downarrow$ & \textbf{31.0\%}$\downarrow$ & -10.1\% & \textbf{53.3\%}$\downarrow$ & \textbf{46.0\%}$\downarrow$ \\
\bottomrule
\end{tabular*}
\caption{Task success and resource use. SR denotes the success rate (\%). Alone uses only Jev. Rank and Requery apply loop-breaking controls without LLM assistance. Full uses the LLM policy throughout. Routed uses Jev + \method{}. Time and cost are per-task means and include both policies for routed runs. The bottom rows average the seven LLM comparisons equally. Mean gain gives the signed success-rate difference (Routed minus Full) and the relative time/cost reductions ($\downarrow$). The success-rate difference is expressed on the percentage scale, not as relative growth. Resource reductions are computed per pair before averaging. Cost accounting is in Appendix~\ref{app:full}.}
\label{tab:main}
\end{table*}

\paragraph{Policies.}
Jev~\citep{almeida2026jev} is our main System~1 decision model. We also evaluate StartLux-4B~\citep{startlux2026decision}, Laya~\citep{laya2026}, JEV-27B-VL~\citep{autotrust2026jevvl}, Kev-4B~\citep{palmer2026kev4b}, and OneJev1.6-4B~\citep{onejev2026}. All are frozen and receive text only. For LLM assistance, we evaluate \llmlogo{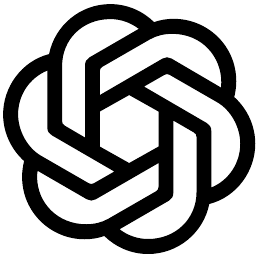}GPT-6-Astra~\citep{openai2026gpt6astra}, \llmlogo{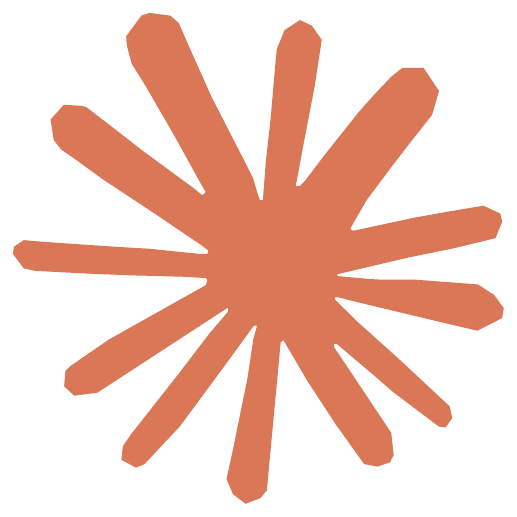}Claude Opus 5~\citep{anthropic2026opus5}, \llmlogo{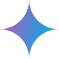}Gemini 3.8 Flash~\citep{google2026gemini38flash}, \llmlogo{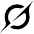}Grok 4.5~\citep{xai2026grok45}, \llmlogo{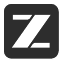}GLM-5.3~\citep{zai2026glm53}, \llmlogo{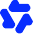}Qwen3.8-Flash~\citep{qwen2026flash38}, and \llmlogo{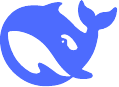}DeepSeek-V4.1-Flash~\citep{deepseek2026v41}. Each uses ReAct~\citep{yao2023react} on ALFWorld and a state-tracking prompt adapted from StateAct~\citep{rozanov2025stateact} on MapReader and cooking. These prompts elicit reasoning and task-state tracking to support subgoal planning and action selection. Model-generation settings are given in Appendix~\ref{app:impl}. The main comparisons use $\tau=2$ and $B=2$. The threshold and takeover-budget analyses and six-model evaluation use DeepSeek-V4.1-Flash. The random-intervention comparison uses GLM-5.3 with a uniform output limit (Appendix~\ref{app:random}).

\paragraph{Baselines.}
We compare standalone Jev, the complete ReAct or StateAct policy for each LLM, and two loop-breaking controls. \emph{Rank} replaces a repeated proposal with Jev's highest-scoring action below the repetition threshold. \emph{Requery} masks actions at the threshold and queries Jev again. Both use the same repetition check as \method{} but never call the LLM policy. All conditions using a decision model receive the same task prompt, fact table, and full history.

\paragraph{Metrics.}
We report task success, mean time per task, and mean USD cost per task. Time sums the measured durations of Jev and LLM requests within each task before averaging. Routed totals include both policies. Comparisons use the same task IDs for success and resources. Accounting details and success counts appear in Appendix~\ref{app:full}. Request savings, $1-C_{\method{}}/C_{\pi_\ell}$, compare routed and complete execution with the same LLM (Figure~\ref{fig:main-results}).

\subsection{Main Results}
\label{sec:exp-main}
Table~\ref{tab:main} compares standalone Jev, full LLM control, and routed execution.

\begin{figure*}[t]
\centering
\includegraphics[width=\textwidth]{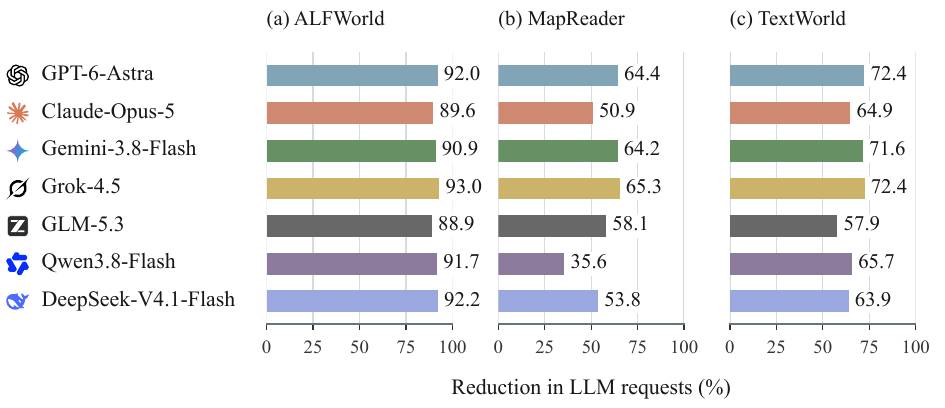}
\caption{Reduction in LLM requests under \method{} across seven LLMs. Each bar compares routed and complete policies using the same LLM and matched task sets, on a common 0--100\% scale. Task success, per-task time, and USD cost are in Table~\ref{tab:main}. The success counts and mean episode lengths are in Appendix~\ref{app:full}.}
\label{fig:main-results}
\end{figure*}

\paragraph{Coordinating success, time, and cost.}
The comparisons connect task-level gains to control allocation. Standalone Jev and routed runs use the same context construction, so their success gap reflects adding LLM assistance on this common decision interface. The loop signal transfers control to an LLM policy that uses feedback for task-level planning and reasoning. Bounded takeovers then return action selection to Jev. This division preserves fast decisions between interventions while using LLM planning to restore progress. On ALFWorld and MapReader, mean success rates are only 0.2 and 0.6 percentage points below those under full LLM control, respectively, while LLM requests are substantially reduced (Figure~\ref{fig:main-results}). In 20 of the 21 LLM--benchmark comparisons, both per-task time and cost decrease. Across benchmarks, mean paired reductions range from 30.7--66.8\% for time and 31.0--71.0\% for cost, demonstrating resource savings over complete task execution. By coordinating when each policy controls execution, the routing strategy thus successfully converts local decision efficiency into task-level gains.

\paragraph{Controls without LLM assistance.}
\label{sec:exp-break}
Rank and Requery improve over standalone Jev but recover fewer tasks than routing (Table~\ref{tab:main}). With DeepSeek-V4.1-Flash, \method{} solves 8, 35, and 17 more tasks than the better control on ALFWorld, MapReader, and cooking. On MapReader, Rank and Requery reduce executions beyond the repetition threshold from 5,462 to 42 and 37, fewer than the 282 under \method{}, yet both reach only 62\% success. Their improvement shows that acting on repetition helps stalled tasks progress. However, simply preventing repeated actions does not ensure a sequence of actions that advances the task. The further gain under \method{} comes from handing action selection to an LLM policy. The LLM uses the remaining goal and feedback to plan actions that break the loop before control returns to Jev. The loop signal thus coordinates the strengths of the two model types: Jev specializes in local decisions, while LLM assistance supports task-level planning and recovery.

\subsection{Analysis of Loop-Triggered Coordination}
\label{sec:analysis}
\paragraph{Repetition and task failure.}
We first examine whether the observed loop signal marks a gap between continued local action selection and task progress. In standalone Jev runs, 30 ALFWorld, 61 MapReader, and 43 cooking tasks enter state--action loops and subsequently exhaust the 100-step budget. Repetition denotes executing an action at least a third time under the same observed state. In every MapReader and cooking failure at the step limit, at least half of the final 20 actions exceed this threshold, showing that repetition continues to consume the remaining budget. On MapReader, Jev solves only 3 of 64 tasks with repetition, versus all 36 without it. We next compare against random intervention before examining when assistance should begin and how long it should last.

\paragraph{Comparison with random intervention.}
\label{sec:exp-random}
To test whether assistance volume explains the gains, we compare \method{} with random intervention using the same Jev-1.13.0 model and GLM-5.3 policy. Both use the same prompt framework and a 4,096-token output limit. The random policy selects the LLM independently at each step with a fixed benchmark-level probability and returns control after one action. \method{} instead uses loop-triggered takeovers of at most two actions. Probabilities are calibrated using LLM action counts alone. Appendix~\ref{app:random} documents all calibration and evaluation rounds.

\begin{table}[t]
\centering
\fontsize{9.5}{11.5}\selectfont
\setlength{\tabcolsep}{3pt}
\begin{tabular*}{\columnwidth}{@{\extracolsep{\fill}}lrrr@{}}
\toprule
Policy & SR (\%) & LLM & Env. \\
\midrule
\multicolumn{4}{@{}l}{ALFWorld} \\
\method{} & 99.3 & 0.98 & 28.16 \\
Random (lower) & 96.3 & 0.81 & 27.46 \\
Random (higher) & 97.8 & 1.15 & 23.33 \\
\midrule
\multicolumn{4}{@{}l}{MapReader} \\
\method{} & 99.0 & 4.50 & 21.97 \\
Random & 74.0 & 4.63 & 43.98 \\
\midrule
\multicolumn{4}{@{}l}{TextWorld} \\
\method{} & 86.0 & 4.20 & 23.35 \\
Random & 75.0 & 4.21 & 35.61 \\
\bottomrule
\end{tabular*}
\caption{Comparison with random intervention. LLM and Env. are executed LLM and total environment actions per task. MapReader and TextWorld random use is within 10\% of routed LLM actions. Both ALFWorld random rounds fall outside this tolerance. All rounds appear in Appendix~\ref{app:random}.}
\label{tab:random-strategy}
\end{table}

With closely matched LLM action counts, \method{} improves success by 25 and 11 percentage points on MapReader and TextWorld (Table~\ref{tab:random-strategy}). It also reduces mean environment actions by 50.0\% and 34.4\%, respectively. Assistance volume alone therefore does not explain the gains. The results support coupling execution feedback with bounded, consecutive assistance to coordinate local decisions and task-level recovery. This comparison tests the complete policy, including intervention timing and takeover continuity, rather than the trigger alone. ALFWorld shows smaller success differences, and neither additional random round meets the budget tolerance.

\subsubsection{Choice of the Repetition Threshold}
\label{sec:exp-threshold}
\begin{table}[t]
\centering
\fontsize{9.5}{11.5}\selectfont
\setlength{\tabcolsep}{2pt}
\begin{tabular*}{\columnwidth}{@{\extracolsep{\fill}}crrrrrr@{}}
\toprule
& \multicolumn{2}{c}{ALFWorld} & \multicolumn{2}{c}{MapReader} & \multicolumn{2}{c}{TextWorld} \\
\cmidrule(lr){2-3}\cmidrule(lr){4-5}\cmidrule(l){6-7}
$\tau$ & SR (\%) & Req. & SR (\%) & Req. & SR (\%) & Req. \\
\midrule
1 & 100.0 & 32 & 80.0 & 718 & 85.0 & 256 \\
2 & 100.0 & 14 & 80.0 & 538 & 85.0 & 281 \\
3 & 95.0 & 22 & 60.0 & 640 & 80.0 & 313 \\
\bottomrule
\end{tabular*}
\caption{Threshold sensitivity with Jev and DeepSeek-V4.1-Flash on 20 additional tasks per benchmark. SR is task success. Req. counts all LLM HTTP requests. The takeover budget remains $B=2$.}
\label{tab:threshold}
\end{table}

\begin{table}[!tb]
\centering
\fontsize{9.5}{11.5}\selectfont
\setlength{\tabcolsep}{2pt}
\begin{tabular*}{\columnwidth}{@{\extracolsep{\fill}}crrrrrr@{}}
\toprule
& \multicolumn{2}{c}{ALFWorld} & \multicolumn{2}{c}{MapReader} & \multicolumn{2}{c}{TextWorld} \\
\cmidrule(lr){2-3}\cmidrule(lr){4-5}\cmidrule(l){6-7}
$B$ & SR (\%) & Req. & SR (\%) & Req. & SR (\%) & Req. \\
\midrule
1 & 100.0 & 13 & 75.0 & 645 & 85.0 & 241 \\
2 & 100.0 & 14 & 80.0 & 538 & 85.0 & 281 \\
3 & 95.0 & 52 & 75.0 & 615 & 80.0 & 243 \\
\bottomrule
\end{tabular*}
\caption{Takeover-budget sensitivity on the same 20 tasks per benchmark as Table~\ref{tab:threshold}, with Jev, DeepSeek-V4.1-Flash, and $\tau=2$. SR is task success. Req. counts all LLM HTTP requests. The $B=2$ trajectories are reused from that study.}
\label{tab:takeover}
\end{table}

\begin{table*}[t]
\centering
\fontsize{9.5}{11.5}\selectfont
\setlength{\tabcolsep}{2.15pt}
\begin{tabular*}{\textwidth}{@{\extracolsep{\fill}}lrrr@{\hspace{9pt}}rrr@{\hspace{9pt}}rrr@{}}
\toprule
& \multicolumn{3}{c}{ALFWorld} & \multicolumn{3}{c}{MapReader} & \multicolumn{3}{c}{Cooking} \\
\cmidrule(lr){2-4}\cmidrule(lr){5-7}\cmidrule(l){8-10}
Decision model & Alone (\%) & Routed (\%) & Gain & Alone (\%) & Routed (\%) & Gain & Alone (\%) & Routed (\%) & Gain \\
\midrule
Jev & 77.6 & 94.8 & \textcolor{routeblue}{\textbf{+17.2\%}} & 39.0 & 97.0 & \textcolor{routeblue}{\textbf{+58.0\%}} & 54.0 & 83.0 & \textcolor{routeblue}{\textbf{+29.0\%}} \\
JEV-27B-VL & 85.1 & 97.0 & \textcolor{routeblue}{\textbf{+11.9\%}} & 67.0 & 99.0 & \textcolor{routeblue}{\textbf{+32.0\%}} & 74.0 & 84.0 & \textcolor{routeblue}{\textbf{+10.0\%}} \\
StartLux-4B & 53.7 & 93.3 & \textcolor{routeblue}{\textbf{+39.6\%}} & 32.0 & 91.0 & \textcolor{routeblue}{\textbf{+59.0\%}} & 28.0 & 43.0 & \textcolor{routeblue}{\textbf{+15.0\%}} \\
OneJev1.6-4B & 20.9 & 93.3 & \textcolor{routeblue}{\textbf{+72.4\%}} & 28.0 & 98.0 & \textcolor{routeblue}{\textbf{+70.0\%}} & 14.0 & 23.0 & \textcolor{routeblue}{\textbf{+9.0\%}} \\
Kev-4B & 17.9 & 94.8 & \textcolor{routeblue}{\textbf{+76.9\%}} & 17.0 & 99.0 & \textcolor{routeblue}{\textbf{+82.0\%}} & 17.0 & 50.0 & \textcolor{routeblue}{\textbf{+33.0\%}} \\
Laya & 0.0 & 95.5 & \textcolor{routeblue}{\textbf{+95.5\%}} & 0.0 & 88.0 & \textcolor{routeblue}{\textbf{+88.0\%}} & -- & -- & -- \\
\bottomrule
\end{tabular*}
\caption{Transfer across six frozen decision models with unchanged $\tau=2$ and $B=2$, using DeepSeek-V4.1-Flash for LLM assistance. Alone and Routed give success rates in percent. Gain is their signed absolute difference (Routed minus Alone), not relative growth. The complete LLM policy achieves 94.0\% on ALFWorld, 98.0\% on MapReader, and 92.0\% on cooking. Resource results are in Appendix~\ref{app:full}. Laya cooking results are unavailable.}
\label{tab:transfer}
\end{table*}

The repetition threshold balances LLM assistance against continued local execution. We compare $\tau\in\{1,2,3\}$ on 20 fixed tasks per benchmark, disjoint from the main evaluation sets. The LLM policy, prompts, counter updates, two-step takeover budget, and 100-action limit remain fixed. Appendix~\ref{app:impl} gives the sampling protocol.

At $\tau=2$, success matches $\tau=1$ on each benchmark with 17.2\% fewer total LLM requests (833 versus 1,006). Earlier handoff therefore adds request overhead without improving observed success. Raising the threshold to $\tau=3$ lowers success and increases requests on all three benchmarks (Table~\ref{tab:threshold}). Delaying the trigger therefore does not necessarily save requests. This pattern is consistent with delayed assistance increasing later recovery demand. Intervention timing should thus be assessed through task-level success and total request use. We retain $\tau=2$ as a shared compromise. Although $\tau=1$ uses fewer requests on TextWorld, $\tau=2$ provides the better aggregate trade-off without benchmark-specific thresholds.

\subsubsection{Choice of the Takeover Budget}
\label{sec:exp-takeover}

The threshold determines when assistance begins. The takeover budget determines when Jev resumes control. Table~\ref{tab:takeover} reports $B\in\{1,2,3\}$ on the same 60 additional tasks, with $\tau=2$ and all other settings fixed. Each LLM action receives fresh feedback. The reuse of the $B=2$ runs is specified in Appendix~\ref{app:impl}.

Across all 60 tasks, two-step takeovers achieve the highest observed success and the fewest LLM requests among these three settings: 53/60 successes with 833 requests, compared with 52/60 and 899 at $B=1$, and 50/60 and 910 at $B=3$. Requests fall by 7.3\% and 8.5\%, respectively. One-step assistance produces 887 takeover segments, versus 412 at $B=2$. Thus shorter interventions can require more recurrent assistance over a complete task. Three steps also fail to improve aggregate success or request use. These results support two steps as a shared default based on task-level success and total LLM demand.

\subsection{Transfer across Decision Models}
\label{sec:exp-cost}

A coordination signal derived from execution should remain usable when the default selector changes. Table~\ref{tab:transfer} tests this requirement with the same repetition threshold and takeover budget across six frozen decision models, without refitting the router. Standalone success varies from 0--85.1\% on ALFWorld and 0--67.0\% on MapReader. Routing narrows these ranges to 93.3--97.0\% and 88.0--99.0\%, respectively. This improvement spans models with widely different standalone capabilities and requires no model-specific routing threshold.

The gains persist across markedly different standalone capabilities. Kev-4B improves by 76.9, 82.0, and 33.0 percentage points on ALFWorld, MapReader, and cooking, respectively. Cooking gains are positive for all five evaluated decision models, although their routed success remains more variable. These results support the transferability of the execution-based routing rule without fitting model-specific thresholds. Request use, per-task time, and available cost estimates are reported separately in Appendix~\ref{app:full}.

\subsection{Irreversible Errors in Cooking}
\label{sec:exp-boundary}
TextWorld cooking exposes the boundary of repetition-triggered routing. With DeepSeek-V4.1-Flash, \method{} reduces failures at the step limit from 43 to 2, but 11 games still end in a loss after irreversible actions (Table~\ref{tab:failures}). More capable LLMs sharpen this diagnosis: GPT-6-Astra and Gemini 3.8 Flash each solve 100/100 games under complete control but 83/100 when routed. All 17 routed failures for each LLM terminate after an irreversible Jev action, and none reaches the step limit. Grok 4.5 shows the same pattern: all 13 routed failures end in such losses. These results locate the remaining gap in task-destroying actions that repetition-triggered recovery cannot undo. Preventing those actions requires a signal available before irreversible execution.

\begin{table}[t]
\centering
\fontsize{10}{12}\selectfont
\setlength{\tabcolsep}{4pt}
\begin{tabular*}{\columnwidth}{@{\extracolsep{\fill}}lrrr@{}}
\toprule
Method & Success & Lost & Limit \\
\midrule
Jev & 54 & 3 & 43 \\
\quad + Rank & 66 & 13 & 21 \\
\quad + \method{} & 83 & 11 & 2 \\
DeepSeek + StateAct & 92 & 6 & 0 \\
\bottomrule
\end{tabular*}
\caption{Success and selected failure outcomes in 100 cooking games. Routed and StateAct runs use DeepSeek-V4.1-Flash. Lost denotes a game loss after an irreversible action. Limit denotes exhaustion of the action budget. Requery matches Rank.}
\label{tab:failures}
\end{table}

\section{Conclusion}
\label{sec:conclusion}
\method{} uses observed execution loops to allocate control between fast, low-cost decision models and LLM policies that reason about task-level planning and recovery. Short, feedback-driven takeovers improve Jev's task success, while reducing per-task time and cost in 20 of 21 LLM--benchmark comparisons. Threshold and takeover-budget comparisons support the shared defaults, while loop-breaking controls show that suppressing repetition alone is insufficient. The unchanged rule improves six frozen decision models. Irreversible actions expose its boundary: recovery cannot undo a task-destroying error that precedes the signal. Execution-based routing therefore offers a practical way to turn local decision efficiency into task-level gains by coordinating complementary policies.
\section*{Limitations}
\method{} uses a fixed two-step takeover budget and returns control without assessing whether the decision model can sustain progress. Future work could make the handover depend on task progress.

Repetition-triggered assistance cannot prevent a single irreversible error before a loop appears. In cooking, destroying a required ingredient can make task completion impossible despite later LLM assistance. Checks before irreversible actions are a possible direction for extending the method.

The current implementation uses text observations. Extending the routing signal to visual and other multimodal observations remains future work.

\bibliography{sample}

\appendix
\raggedbottom
\renewcommand{\dbltopfraction}{0.95}
\renewcommand{\textfraction}{0.03}
\section{Extended Results and Resource Accounting}
\label{app:full}

\paragraph{Resource accounting.}
API time sums recorded Jev and LLM request durations per task, then averages over matched task IDs. It excludes client-side queue waits and is distinct from end-to-end wall-clock time. GPT-6-Astra, Claude Opus 5, Gemini 3.8 Flash, Grok 4.5, and GLM-5.3 use API-returned USD charges. Qwen and DeepSeek use token-based estimates. Per million tokens, DeepSeek costs USD~0.30 for uncached input, USD~0.006 for cached input, and USD~1.20 for output. Qwen costs CNY~0.80 for input and CNY~2.70 for output, converted at CNY~6.6723 per USD (2026-11-04), without off-peak or additional cache discounts. Jev input tokens cost USD~0.042 per million, with no output charge. Routed totals include both policies.

\paragraph{Checks on resource savings.}
On the descriptive subset of 83 cooking tasks solved by both GPT-6-Astra policies, routing reduces cumulative API time by 68.2\% and USD cost by 74.7\%, including Jev. Savings therefore persist when both policies succeed. Excluding GLM-5.3, whose output-budget procedure is described in Appendix~\ref{app:impl}, both resources still decrease in 17 of 18 comparisons. Equal-weight mean paired reductions remain positive on each benchmark.

\begin{table*}[t]
\centering
\fontsize{9.5}{11.5}\selectfont
\setlength{\tabcolsep}{4pt}
\begin{tabular*}{\textwidth}{@{\extracolsep{\fill}}lrrrr@{}}
\toprule
Decision model / policy & LLM requests & Saved (\%) & Time (s) & Saved (\%) \\
\midrule
\multicolumn{5}{@{}l}{\textit{ALFWorld (134 tasks)}} \\
Full LLM & 2,941 & -- & 157.3 & -- \\
Jev & 230 & 92.2 & 23.6 & 85.0 \\
JEV-27B-VL & 176 & 94.0 & 47.6 & 69.8 \\
StartLux-4B & 419 & -- & 70.8 & -- \\
OneJev1.6-4B & 678 & 76.9 & 163.4 & -3.8 \\
Kev-4B & 895 & 69.6 & 121.1 & 23.0 \\
Laya & 2,199 & -- & 216.7 & -- \\
\midrule
\multicolumn{5}{@{}l}{\textit{MapReader (100 tasks)}} \\
Full LLM & 1,740 & -- & 193.0 & -- \\
Jev & 804 & 53.8 & 102.4 & 47.0 \\
JEV-27B-VL & 433 & 75.1 & 50.3 & 74.0 \\
StartLux-4B & 1,016 & 41.6 & 204.2 & -5.8 \\
OneJev1.6-4B & 1,346 & 22.6 & 155.9 & 19.3 \\
Kev-4B & 1,413 & 18.8 & 146.4 & 24.2 \\
Laya & 1,763 & -1.3 & 145.4 & 24.7 \\
\midrule
\multicolumn{5}{@{}l}{\textit{Cooking (100 tasks)}} \\
Full LLM & 1,989 & -- & 198.0 & -- \\
Jev & 718 & 63.9 & 92.4 & 53.4 \\
JEV-27B-VL & 408 & 79.5 & 64.2 & 67.6 \\
StartLux-4B & 795 & 60.0 & 117.8 & 40.5 \\
OneJev1.6-4B & 641 & 67.8 & 114.0 & 42.4 \\
Kev-4B & 1,083 & 45.6 & 140.5 & 29.1 \\
\bottomrule
\end{tabular*}
\caption{Resource use for the transfer runs in Table~\ref{tab:transfer}. Full LLM is the complete DeepSeek-V4.1-Flash policy. Saved is the relative reduction against a matching reference, with negative values indicating increases. Time is the per-task sum for both policies. Dashes mark unmatched comparisons. Jev costs are already reported in Table~\ref{tab:main}. Other selectors lack consistently priced usage for both policies.}
\label{tab:full}
\end{table*}

\begin{table*}[t]
\centering
\fontsize{9.5}{11.5}\selectfont
\setlength{\tabcolsep}{3pt}
\begin{tabular*}{\textwidth}{@{\extracolsep{\fill}}lrrr@{\hspace{8pt}}rrr@{\hspace{8pt}}rrr@{}}
\toprule
& \multicolumn{3}{c}{ALFWorld} & \multicolumn{3}{c}{MapReader} & \multicolumn{3}{c}{Cooking} \\
\cmidrule(lr){2-4}\cmidrule(lr){5-7}\cmidrule(l){8-10}
LLM & Full & Routed & Steps F/R & Full & Routed & Steps F/R & Full & Routed & Steps F/R \\
\midrule
GPT-6-Astra & 134/134 & 134/134 & 9.5/27.1 & 100/100 & 100/100 & 12.7/22.1 & 100/100 & 83/100 & 15.0/22.3 \\
Claude Opus 5 & 134/134 & 132/134 & 11.0/28.6 & 100/100 & 99/100 & 13.0/25.4 & 91/100 & 81/100 & 16.5/25.0 \\
Gemini 3.8 Flash & 134/134 & 134/134 & 8.2/28.2 & 100/100 & 100/100 & 12.3/21.9 & 100/100 & 83/100 & 14.4/22.6 \\
Grok 4.5 & 134/134 & 134/134 & 11.7/27.5 & 98/98 & 98/98 & 12.3/21.7 & 100/100 & 87/100 & 14.4/22.6 \\
GLM-5.3 & 134/134 & 133/134 & 10.2/29.5 & 100/100 & 100/100 & 12.8/23.6 & 95/100 & 89/100 & 15.2/25.4 \\
Qwen3.8-Flash & 131/134 & 131/134 & 14.5/29.7 & 100/100 & 98/100 & 13.3/25.2 & 82/100 & 83/100 & 23.4/27.7 \\
DeepSeek-V4.1-Flash & 126/134 & 127/134 & 21.2/31.0 & 98/100 & 97/100 & 16.1/25.5 & 92/100 & 83/100 & 18.4/24.4 \\
\bottomrule
\end{tabular*}
\caption{Success counts and mean episode lengths for Table~\ref{tab:main}. Full and Routed report successes over the evaluated task count. Steps F/R gives mean environment actions for complete and routed policies, including unsuccessful episodes. Request comparisons use the same task IDs as these counts.}
\label{tab:lengths}
\end{table*}

Resource use varies across decision models. Relative savings in Table~\ref{tab:full} require a matching complete-policy configuration. The ALFWorld evaluations of Laya and StartLux-4B therefore report absolute use only. Table~\ref{tab:lengths} provides the complete success counts and episode lengths for the main comparisons.

\section{Confidence-Based Routing}
\label{app:conf}
\label{sec:exp-confidence}
We compare repetition-only routing with an additional confidence gate on the same 134 ALFWorld tasks. The gate requests assistance when the API-reported confidence $c_t<0.15$ or the repetition trigger fires. This confidence field is not calibrated as a probability of action correctness. Both policies use the same context, LLM, and two-step takeover budget.

\begin{table}[tbp]
\centering
\fontsize{10}{12}\selectfont
\setlength{\tabcolsep}{4pt}
\begin{tabular*}{\columnwidth}{@{\extracolsep{\fill}}lrr@{}}
\toprule
Routing rule & Success & Requests \\
\midrule
Repetition only (\method{}) & 127/134 & 230 \\
\quad + confidence $<0.15$ & 127/134 & 399 \\
\bottomrule
\end{tabular*}
\caption{Confidence-gate comparison on 134 matched ALFWorld tasks. Requests include HTTP retries.}
\label{tab:conf}
\end{table}

The gate increases requests by 73.5\% without improving aggregate success. The solved sets differ by one task in each direction. On this comparison, confidence adds overhead without a net success gain.

\section{Implementation Details}
\label{app:impl}
\paragraph{State keys.}
The keys hash a canonical JSON representation of selected public fields. They exclude step counters and growing execution histories. ALFWorld uses its confirmed-fact dictionary and sorted admissible commands. Its fact dictionary records location, carrying status, object locations and verified conditions, container states, inspected contents, and observed device activations. MapReader uses the current action response, room description, and inventory. TextWorld uses the current room description, inventory, and candidate-ID-to-command mapping. The fact table remains available to both policies even when its fields are not included in the key.
\paragraph{Counter updates.}
On ALFWorld, counters accumulate decision-model executions until new public information is confirmed or a takeover ends. LLM executions do not increment these counters. New information includes a first object or receptacle discovery, a verified processing operation, a first required placement, or an observed device activation. On MapReader and TextWorld, both policies' executed actions increment the corresponding state--action count, and counts persist for the episode. No counter is incremented for an intercepted proposal. Across environments, we therefore transfer the trigger-and-bounded-takeover framework with public-state adapters, rather than an identical counter implementation. In the six-model transfer study, each benchmark's adapter, threshold, and takeover budget remain fixed across decision models.
\paragraph{Takeovers.} An intercepted proposal is discarded and does not enter the history. Environment termination ends a takeover immediately.
\paragraph{Request counting.} LLM requests count every call sent to the policy, including retries and repeated calls for the same decision. Executed environment actions are counted separately.
\paragraph{Threshold sensitivity protocol.}
The comparison in \S\ref{sec:exp-threshold} uses Jev-1.13.0 and DeepSeek-V4.1-Flash on 20 fixed tasks per benchmark: ALFWorld \texttt{valid\_seen} (4/4/3/3/3/3 across task families), TextWorld validation (four games per level, 5--9), and MapReader (new test seeds, 30 locations, maximum distance 8). SHA-256 ordering selects tasks without overlap with the main and prior exploratory sets. Thresholds run independently with unchanged prompts, state keys, counters, and generation settings. These sensitivity subsets were evaluated after the main experiments.

\paragraph{Takeover-budget sensitivity protocol.}
Table~\ref{tab:takeover} uses the same 60 tasks at $\tau=2$. The 60 $B=2$ trajectories are reused from the threshold study, including failures. The $B=1,3$ comparisons add 120 trajectories under the same settings. Each task runs once per budget, following \S\ref{sec:how}.
\paragraph{Repeated takeover failures.}
The seven unsuccessful ALFWorld runs with Jev and DeepSeek-V4.1-Flash reach the step limit after 8--13 takeovers. The records do not distinguish assistance that fails to establish progress from progress lost after handoff.
\paragraph{Prompt interfaces.}
ALFWorld uses two task-family ReAct demonstrations followed by the shared current context and all admissible commands. The output instruction requests a brief \texttt{Thought} (at most 25 words) and one \texttt{Action:} line with an exact command. MapReader and TextWorld use state-tracking prompts that ask for the goal, visited and current locations, inventory, remaining requirements, a thought, and one candidate action. Each selected action is executed before fresh feedback is supplied. The generated state summary is not treated as a verified fact.
\paragraph{LLM generation settings.}
All LLM policies use temperature 0 with ReAct or StateAct prompts. Qwen and DeepSeek disable the API's separate reasoning mode. GPT-6-Astra, Claude Opus 5, Gemini 3.8 Flash, and Grok 4.5 use output limits of 1,024 tokens on ALFWorld and 2,048 on MapReader and cooking. GLM-5.3 initially uses 192/640/640 tokens. The same output-error rule reruns 14 affected Full/Routed trajectories with expanded limits of 1,024/2,048/2,048 and a 2,048-token format-retry limit. Success and episode length use final trajectories. Requests, API time, and costs include original and rerun usage once.
\paragraph{Evaluation details.} Cooking reuses 20 root seeds across five difficulty levels. The main ALFWorld split was used in earlier exploration, and some cooking games informed prompt development. These results are therefore fixed-benchmark evaluations rather than untouched development-free tests. In the ALFWorld context comparison, Jev solves 14/134 tasks with only the goal and current observation, versus 104/134 with the structured input. This comparison adds the context fields jointly.

\paragraph{Reproduction materials.}
The accompanying \texttt{reproducibility/} directory contains prompt templates, ordered ALFWorld demonstrations, task IDs and game hashes, the 60-task sensitivity manifest, generation settings, and counter specifications.

\section{Motivation Evidence and Loop Accounting}
\label{app:diagnostics}
\paragraph{Static tool selection.}
Figure~\ref{fig:motivation} contrasts local tool selection with interactive task completion. We use 200 BFCL Multiple cases~\citep{patil2025bfcl}, each with two to four documented tools plus a no-call option. Jev-1.13.0 selects a tool ID from the user request and complete tool documentation, correctly selecting all 200. This measures selection only, without generating arguments or executing tools. It is not a complete BFCL score.

\paragraph{Loop measurements.}
We recompute the repetition counts on standalone Jev trajectories using the keys and counter updates in Appendix~\ref{app:impl}. A repeated step is an executed state--action pair whose pre-execution count is already at least two. The first two executions are not counted. A loop onset is the first such step for a pair within a counting interval (pre-execution count exactly two). This counts distinct threshold crossings, not independent cycles: an alternating cycle may yield an onset for each of its two actions. Mean onsets divides their total by all evaluated tasks, including tasks without repetition.

For a task set $\mathcal{D}$, let $R_i$ be the number of repeated steps and $T_i$ its total executed steps. We report the pooled step share
\begin{equation}
\rho(\mathcal{D})=\frac{\sum_{i\in\mathcal{D}}R_i}{\sum_{i\in\mathcal{D}}T_i}.
\label{eq:loop-share}
\end{equation}
\begin{table}[tbp]
\centering
\fontsize{9.5}{11.5}\selectfont
\setlength{\tabcolsep}{3pt}
\begin{tabular*}{\columnwidth}{@{\extracolsep{\fill}}lrrr@{}}
\toprule
& Mean & \multicolumn{2}{c}{Repeated-step share (\%)} \\
\cmidrule(l){3-4}
Benchmark & onsets & All tasks & Failed tasks \\
\midrule
ALFWorld & 1.522 & 28.26 & 47.63 \\
MapReader & 2.370 & 83.52 & 87.87 \\
TextWorld & 2.090 & 67.96 & 77.46 \\
\bottomrule
\end{tabular*}
\caption{Standalone Jev repetition. ALFWorld, MapReader, and TextWorld contain 134, 100, and 100 tasks, of which 30, 61, and 46 fail. Failed-task shares are respectively 1,429/3,000, 5,360/6,100, and 3,375/4,357 steps.}
\label{tab:loop-accounting}
\end{table}

The failed-task share uses the same ratio restricted to unsuccessful episodes, rather than averaging per-task percentages. Table~\ref{tab:loop-accounting} reports both shares for the same standalone runs as Table~\ref{tab:main}.

\paragraph{Repetition and failure.}
Repeated steps occur in 44/134 ALFWorld, 64/100 MapReader, and 51/100 cooking tasks, of which 30, 61, and 44 fail. Repetition also occurs in successful trajectories, so we use it as an assistance signal rather than a deterministic failure label.

\section{Random-Intervention Comparison}
\label{app:random}
\paragraph{Settings and random policy.}
Jev-1.13.0 and GLM-5.3 use the same 134/100/100 main task IDs, benchmark-specific prompts, context, and 100-action limit. Both policies use a 4,096-token output limit. These are separate runs from the historical GLM comparisons in Table~\ref{tab:main}. Before each action, the random policy independently selects the LLM with a benchmark-level probability $p$, then returns control after one action. \method{} retains $\tau=2$ and at most two consecutive LLM actions.

\paragraph{Resource matching.}
Initial probabilities are calibrated on 12 disjoint ALFWorld, 10 MapReader, and 10 TextWorld tasks using only LLM action counts. Calibration starts from routed LLM actions divided by total environment actions. Up to three rounds select the first probability within 10\% of routed use, or the closest otherwise. The probability is frozen before random evaluation.

If realized evaluation use falls outside tolerance, a prespecified rule permits at most two additional rounds. Each updates $p$ by the ratio of routed to preceding random mean LLM actions, clipped to $[0.0001,0.95]$. The new probability and seed are frozen before that round. The first additional round within tolerance supplies the matched comparison. Table~\ref{tab:random-rounds} retains all rounds and all scheduled tasks, including failures. Selection uses resources, not success. Additional rounds reuse observed evaluation-set resource counts and are descriptive comparisons, rather than independent confirmations.

\begin{table}[t]
\centering
\fontsize{9.5}{11.5}\selectfont
\setlength{\tabcolsep}{2pt}
\begin{tabular*}{\columnwidth}{@{\extracolsep{\fill}}lrrrr@{}}
\toprule
Policy/round & $p$ & Success & LLM & Env. \\
\midrule
\multicolumn{5}{@{}l}{ALFWorld ($N=134$)} \\
\method{} & -- & 133 & 131 & 3,773 \\
Random initial & 0.019396 & 125 & 78 & 3,885 \\
Random extra 1 & 0.032575 & 129 & 109 & 3,680 \\
Random extra 2 & 0.039150 & 131 & 154 & 3,126 \\
\midrule
\multicolumn{5}{@{}l}{MapReader ($N=100$)} \\
\method{} & -- & 99 & 450 & 2,197 \\
Random initial & 0.133259 & 71 & 577 & 4,446 \\
Random extra 1 & 0.103929 & 74 & 463 & 4,398 \\
\midrule
\multicolumn{5}{@{}l}{TextWorld ($N=100$)} \\
\method{} & -- & 86 & 420 & 2,335 \\
Random initial & 0.213666 & 78 & 640 & 3,104 \\
Random extra 1 & 0.140218 & 78 & 475 & 3,329 \\
Random extra 2 & 0.123982 & 75 & 421 & 3,561 \\
\bottomrule
\end{tabular*}
\caption{All random-comparison rounds. Success counts retain the full $N$ for each row. LLM and Env. are total executed actions, not per-task means. Initial evaluation contains 668 trajectories and the additional rounds contain 568.}
\label{tab:random-rounds}
\end{table}

\paragraph{Interpretation.}
MapReader and TextWorld match LLM action use within 10\% . Neither additional ALFWorld round does, so both are reported. Matching actions does not imply matching tokens, latency, or USD cost. The comparison evaluates the complete policy, jointly involving intervention timing and takeover continuity. The accompanying \texttt{random\_intervention.json} records task IDs, probabilities, seeds, calibration rounds, and all aggregate results.

\end{document}